\documentclass[11pt,a4paper]{article}
\usepackage[hyperref]{acl2019}
\aclfinalcopy

\usepackage{times}
\usepackage{amsmath}
\usepackage{amssymb}
\usepackage{tabu}
\usepackage{xcolor}
\usepackage{url}
\usepackage{float}
\usepackage{latexsym}
\usepackage{graphicx}
\usepackage[export]{adjustbox}
\usepackage{makecell}
\usepackage{comment}
\usepackage{tabularx}
\usepackage{subcaption}
\usepackage{caption}
\usepackage[color=blue!20!white]{todonotes}
\usepackage{lscape}
\usepackage{placeins}
\usepackage{siunitx,booktabs}
\usepackage[margin=1in]{geometry}

\usepackage{xspace}

\newcommand{\we}[1]{{e(#1)}}
\newcommand{\wep}[1]{{e'(#1)}}
\newcommand{\enc}{\ensuremath{\mathrm{\textsc{enc}}}}
\newcommand{\gru}{\ensuremath{\mathrm{\textsc{gru}}}\xspace}
\newcommand{\go}{\ensuremath{\mathrm{\textsc{go}}}\xspace}

\newcommand{\lstm}{\ensuremath{\mathrm{\textsc{lstm}}}\xspace}
\newcommand{\rnn}{\ensuremath{\mathrm{\textsc{rnn}}}\xspace}
\newcommand{\avgnv}{\textsc{avg-n2v}\xspace}
\newcommand{\grunv}{\textsc{gru-n2v}\xspace}
\newcommand{\bigru}{\ensuremath{\mathrm{\textsc{bigru}}}\xspace}
\newcommand{\bigrunv}{\textsc{bigru-max-res-n2v}\xspace}
\newcommand{\LR}{\textsc{lr}\xspace}
\newcommand{\CS}{\textsc{cs}\xspace}
\newcommand\nemb{\textsc{ne}\xspace}
\newcommand\deepwalk{\textsc{deepwalk}\xspace}
\newcommand\linemethod{\textsc{line}\xspace}
\newcommand\nodetovec{\textsc{node2vec}\xspace}
\newcommand\wordtovec{\textsc{word2vec}\xspace}
\newcommand\skipgram{\textsc{skipgram}\xspace}
\newcommand\cane{\textsc{cane}\xspace}
\newcommand\cene{\textsc{cene}\xspace}
\newcommand\wane{\textsc{wane}\xspace}
\newcommand\umls{\textsc{umls}\xspace}
\newcommand\gcn{\textsc{gcn}\xspace}

\newcommand\isa{\textsc{is-a}\xspace}
\newcommand\partof{\textsc{part-of}\xspace}
\newcommand\softmax{\textrm{softmax}\xspace}
\newcommand\roc{\textsc{roc}\xspace}
\newcommand\auc{\textsc{auc}\xspace}
\newcommand\fone{\textsc{f1}\xspace}

\title{Embedding Biomedical Ontologies by Jointly Encoding\\
Network Structure and Textual Node Descriptors}

\author{\parbox{12cm}{\centering Sotiris Kotitsas\textsuperscript{1}, Dimitris Pappas\textsuperscript{1,2}, Ion Androutsopoulos\textsuperscript{1}, \vspace{0.5mm} \\
Ryan McDonald\textsuperscript{1,3} \and  Marianna Apidianaki\textsuperscript{4} \vspace{1.5mm}} \\
  \textsuperscript{1}Department of Informatics, Athens University of Economics and Business, Greece \\
  \textsuperscript{2}Institute for Language and Speech Processing, Research Center `Athena', Greece\\
  \textsuperscript{3}Google Research\\
  \textsuperscript{4}CNRS, LLF, Univ. Paris Diderot, France \vspace{1mm} \\
  {\parbox{12cm}{\centering \tt \{p3150077, pappasd, ion\}@aueb.gr  ryanmcd@google.com, marianna@limsi.fr}} \\}

\begin{document}
\maketitle


\begin{abstract}
Network Embedding (\nemb) methods, which map network nodes to low-dimensional feature vectors, have wide applications in network analysis and bioinformatics. Many existing \nemb methods rely only on network structure, overlooking other information associated with the nodes, e.g., text describing the nodes. Recent attempts to combine the two sources of information only consider local network structure. We extend \nodetovec, a well-known \nemb method that considers broader network structure, to also consider textual node descriptors using recurrent neural encoders. Our method is evaluated on link prediction in two networks derived from \umls. Experimental results demonstrate the effectiveness of the proposed approach compared to previous work.
\end{abstract}


\section{Introduction} \label{sec:introduction}

Network Embedding (\nemb) methods map each node of a network to an embedding, meaning a low-dimensional feature vector. They are highly effective in network analysis tasks involving predictions over nodes and edges, for example link prediction~\cite{linkprediction}, and node classification~\cite{vertexclassification}.

\begin{figure}[t]
\centering
\includegraphics[width=0.49\textwidth]{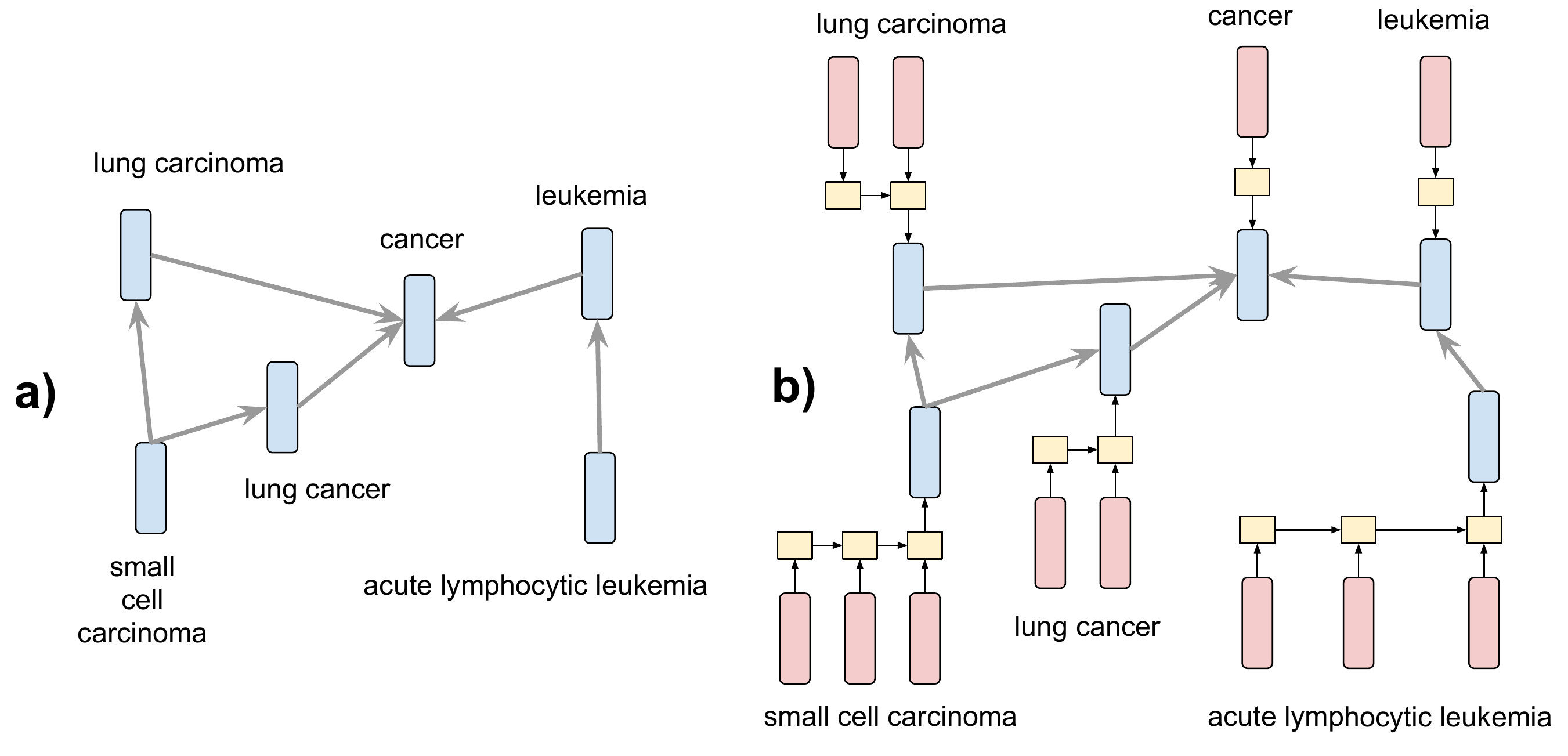}
\vspace{-0.3in}
\caption{Example network with nodes associated with textual descriptors. a) A model where each node is represented by a vector (node embedding) from a look-up table. b) A model where each node embedding is generated compositionally from the word embeddings of its descriptor via an $\rnn$. The latter model can learn node embeddings from both the network structure and the word sequences of the textual descriptors.}
\vspace*{-5mm}
\label{fig:nbl-ex}
\end{figure}

Early \nemb methods, such as \deepwalk \cite{deepwalk}, \linemethod \cite{line}, \nodetovec \cite{node2vec}, \gcn{s} \cite{KipfW16}, leverage information from the network structure to produce embeddings that can reconstruct node neighborhoods. The main advantage of these {\it structure-oriented} methods is that they encode the network context of the nodes, which can be very informative. The downside is that they typically treat each node as an atomic unit, directly mapped to an embedding in a look-up table (Fig.~\ref{fig:nbl-ex}a). There is no attempt to model information other than the network structure, such as textual descriptors (labels) or other meta-data associated with the nodes. 

More recent \nemb methods, e.g., \cane \cite{cane},  \wane \cite{wane}, produce embeddings by combining the network structure and the text associated with the nodes. These {\it content-oriented} methods embed networks whose nodes are rich textual objects (often whole documents). They aim to capture the compositionality and semantic similarities in the text, encoding them with deep learning methods. This approach is illustrated in Fig.~\ref{fig:nbl-ex}b. However, previous methods of this kind considered impoverished network contexts when embedding nodes, usually single-edge hops, as opposed to the non-local structure considered by most structure-oriented methods. 

When embedding biomedical ontologies, it is important to exploit both wider network contexts and textual node descriptors. The benefit of the latter is evident, for example, in `acute leukemia' \isa `leukemia'. To be able to predict (or reconstruct) this \isa relation from the embeddings of `acute leukemia' and `leukemia' (and the word embeddings of their textual descriptors in Fig.~\ref{fig:nbl-ex}b), a \nemb method only needs to model the role of `acute' as a modifier that can be included in the descriptor of a node (e.g., a disease node) to specify a sub-type. This property can be learned (and encoded in the word embedding of `acute') if several similar \isa edges, with `acute' being the only extra word in the descriptor of the sub-type, exist in the network. This strategy would not however be successful in `p53' (a protein) \isa `tumor suppressor', where no word in the descriptors frequently denotes sub-typing. Instead, by considering the broader network context of the nodes (i.e. longer paths that connect them), a \nemb method can detect that the two nodes have common neighbors and, hence, adjust the two node embeddings (and the word embeddings of their descriptors) to be close in the representation space, making it more likely to predict an \isa relation between them.

We propose a new \nemb method that leverages the strengths of both structure and content-oriented approaches. To exploit wide network contexts, we follow \nodetovec \cite{node2vec} and generate random walks to construct the network neighborhood of each node. The \skipgram model~\cite{skipgram} is then used to learn node embeddings that successfully predict the nodes in each walk, from the node at the beginning of the walk. To enrich the node embeddings with information from their textual descriptors, we replace the \nodetovec look-up table with various architectures that operate on the word embeddings of the descriptors. These include simply averaging the word embeddings of a descriptor, and applying recurrent deep learning encoders. The proposed method can  be seen as an extension of \nodetovec that incorporates textual node descriptors. We evaluate several variants of the proposed method on link prediction, a standard evaluation task for \nemb methods. We use two biomedical networks extracted from \umls~\cite{umls}, with \partof and \isa relations, respectively. Our method outperforms several existing structure and content-oriented methods on both datasets. We make our datasets and source code available.\footnote{\url{https://github.com/SotirisKot/Content-Aware-N2V}}


\section{Related work} \label{sec:relatedWork}

Network Embedding (\nemb) methods, a type of representation learning, are highly effective in network analysis tasks involving predictions over nodes and edges. Link prediction has been extensively studied in social networks~\cite{social}, and is particularly relevant to bioinformatics where it can help, for example, to discover interactions between proteins, diseases, and 
genes~\cite{ppi,geneprediction,node2vec}. Node classification can also help analyze large networks by automatically assigning roles or labels to nodes~\cite{role2vec,vertexclassification}. In bioinformatics, this approach has been used to identify proteins whose mutations are linked with particular diseases~\cite{ncdisease}.

A typical structure-oriented \nemb method is \deepwalk~\cite{deepwalk}, which learns node embeddings by applying \wordtovec's \skipgram model~\cite{skipgram} to node sequences generated via random walks on the network. \nodetovec~\cite{node2vec} explores different strategies to perform random walks, introducing hyper-parameters to guide them and generate more flexible neighborhoods. \linemethod \cite{line} learns node embeddings by exploiting first- and second-order proximity information in the network. ~\newcite{sdne} learn node embeddings that preserve the proximity between 2-hop neighbors
using a deep autoencoder.~\newcite{netra} encode node sequences generated via random walks, by mapping the walks to low dimensional embeddings, through an \lstm autoencoder. To avoid overfitting, they use a generative adversarial training process as regularization.
Graph Convolutional Networks (\gcn{s}) are a graph encoding framework that also falls within this paradigm \cite{KipfW16,schlichtkrull2018modeling}. Unlike other methods that use random walks or static neighbourhoods, \gcn{s} use iterative neighbourhood averaging strategies to account for non-local graph structure. All the aforementioned methods
only encode the structural information into node embeddings, ignoring textual or other information that can be associated with the nodes of the network.

Previous work on biomedical ontologies 
(e.g., Gene Ontology, \go) suggested that their terms, which are represented through textual descriptors, have compositional structure. By modeling it, we can create richer representations of the data encoded in the ontologies~\cite{Mungall2004ObolIL, Ogren2003TheCS, Ogren2004ImplicationsOC}.
\newcite{Ogren2003TheCS} strengthen the argument of compositionality by observing that many \go terms contain other \go terms. Also, they argue that substrings 
that are not \go terms appear frequently and often indicate semantic relationships. \newcite{Ogren2004ImplicationsOC} 
use finite state automata to represent \go terms and demonstrate how small conceptual changes can create biologically meaningful candidate terms.

In other work on \nemb methods, \cene~\cite{cene} treats textual descriptors as a special kind of node, and uses bidirectional recurrent neural networks (\rnn{s}) to encode them. \cane~\cite{cane} learns two embeddings per node, a  text-based one and an embedding based on network structure. The text-based one changes when interacting with different neighbors, using a mutual attention mechanism. \wane \cite{wane} also uses two types of node embeddings, text-based and structure-based. For the text-based embeddings, it matches important words across the textual descriptors of different nodes, and aggregates the resulting alignment features. In spite of performance improvements over structure-oriented approaches, these content-aware methods do not thoroughly explore the network structure, since they consider only direct neighbors.

By contrast, we utilize \nodetovec to obtain wider network neighborhoods via random walks, a typical approach of structure-oriented methods, but we also use \rnn{s} to encode the textual descriptors, as in some content-oriented approaches. Unlike \cene, however, we do not treat texts as separate nodes; unlike \cane, we do not learn separate embeddings from texts and network structure; and unlike \wane, we do not align the descriptors of different nodes. We generate the embedding of each node from the word embeddings of its descriptor via the \rnn (Fig.~\ref{fig:nbl-ex}), but the parameters of the \rnn, the word embeddings, hence also the node embeddings are updated during training to predict \nodetovec's neighborhoods. 

Although we use \nodetovec to incorporate network context in the node embeddings, other neighborhood embedding methods, such as \gcn{s}, could easily be used too. Similarly, text encoders other than \rnn{s} could be applied.
For example,~\newcite{Mishra2019AbusiveLD} try to detect abusive language in tweets with a semi-supervised learning approach based on \gcn{s}. They exploit the network structure and also the labels associated with the tweets, taking into account the linguistic behavior of the authors.


\section{Proposed Node Embedding Approach} \label{sec:methods}

Consider a network (graph) $G = \left<V,E,S\right>$, where $V$ is the set of nodes (vertices); $E \subseteq V \times V$ is the set of edges (links) between nodes; and $S$ is a function that maps each node $v \in V$ to its textual descriptor $S(v) = \left<w_{1}, w_{2},\ldots, w_{n}\right>$, where $n$ is the word length of the descriptor, and each word $w_i$ comes from a vocabulary $W$. We consider only undirected, unweighted networks, where all edges represent instances of the same (single) relationship (e.g., \isa or \partof). Our approach, however, can be extended to directed weighted networks with multiple relationship types. We learn an embedding $f(v)\in \mathbb{R}^d$ for each node $v\in V$. As a side effect, we also learn a word embedding $e(w)$ for each vocabulary word $w \in W$.

To incorporate structural information into the node embeddings, we maximize the \emph{predicted} probabilities $p(u|v)$ of observing the \emph{actual} neighbors $u \in N(v)$ of each `focus' node $v \in V$, where $N(v)$ is the neighborhood of $v$, and $p(u|v)$ is predicted from the node embeddings of $u$ and $v$. The neighbors $N(v)$ of $v$ are not necessarily directly connected to $v$. In real-world networks, especially biomedical, many nodes have few direct neighbors. We use \nodetovec~\cite{node2vec} to obtain a larger neighborhood for each node $v$, by generating random walks from $v$. For every focus node $v \in V$, we compute $r$ random walks (paths) $P_{v,i} = \left<v_{i,1}=v, v_{i,2},...,v_{i,k}\right>$ ($i=1, \dots, r$) of fixed length $k$ through the network ($v_{i,j}\in V$).\footnote{Our networks are unweighted, hence we use uniform edge weighting to traverse them. \nodetovec has two hyper-parameters, $p, q$, to control the locality of the walk. We set $p=q=1$ (default values). For efficiency, \nodetovec actually performs $r$ random walks of length $l \geq k$; then it uses $r$ sub-walks of length $k$ that start at each focus node.}
The predicted probability $p(v_{i,j}=u)$ of observing node $u$ at step $j$ of a walk $P_{v,i}$ that starts at focus node $v$ is taken to depend only on the embeddings of $u,v$, i.e., 
$p(v_{i,j}=u) = p(u|v)$, and can be estimated with a \softmax as in the \skipgram model \cite{skipgram}:
\begin{equation}
p(u|v)=
\dfrac{\exp(f'(u) \cdot f(v))}{\sum_{u' \in V} \exp(f'(u') \cdot f(v))} 
\label{eq:softmax}
\end{equation}
where it is assumed that each node $v$ has two different node embeddings, $f(v), f'(v)$, used when $v$ is the focus node or the predicted neighbor, respectively, and $\cdot$ denotes the dot product. \nodetovec minimizes the following objective function:
\begin{equation}
 L = - \sum_{v \in V} \sum_{i=1}^r \sum_{j=2}^k 
        \log p(v_{i,j}|v_{i,1} = v)
\end{equation}
in effect maximizing the likelihood of observing the actual neighbors $v_{i,j}$ of each focus node $v$ that are encountered during the $r$ walks $P_{v,i} = \left<v_{i,1}=v, v_{i,2},...,v_{i,k}\right>$ ($i=1, \dots, r$) from $v$.
Calculating $p(u|v)$ using a softmax (Eq.~\ref{eq:softmax}) is computationally inefficient. We apply negative sampling instead, as in \wordtovec \cite{skipgram}. Thus, \nodetovec is analogous to \skipgram \wordtovec, but using random walks from each focus node, instead of using a context window around each focus word in a corpus.

As already mentioned, the original \nodetovec does not consider the textual descriptors of the nodes. It treats each node embedding $f(v)$ as a vector representing an atomic unit, the node $v$; a look-up table directly maps each node $v$ to its embedding $f(v)$. This does not take advantage of the flexibility and richness of natural language (e.g., synonyms, paraphrases), nor of its compositional nature. To address this limitation, we substitute the look-up table where \nodetovec stores the embedding $f(v)$ of each node $v$ with a neural sequence encoder that produces $f(v)$ from the word embeddings of the descriptor $S(v)$ of $v$.

\begin{figure}[t]
\centering
\includegraphics[width=0.45\textwidth]{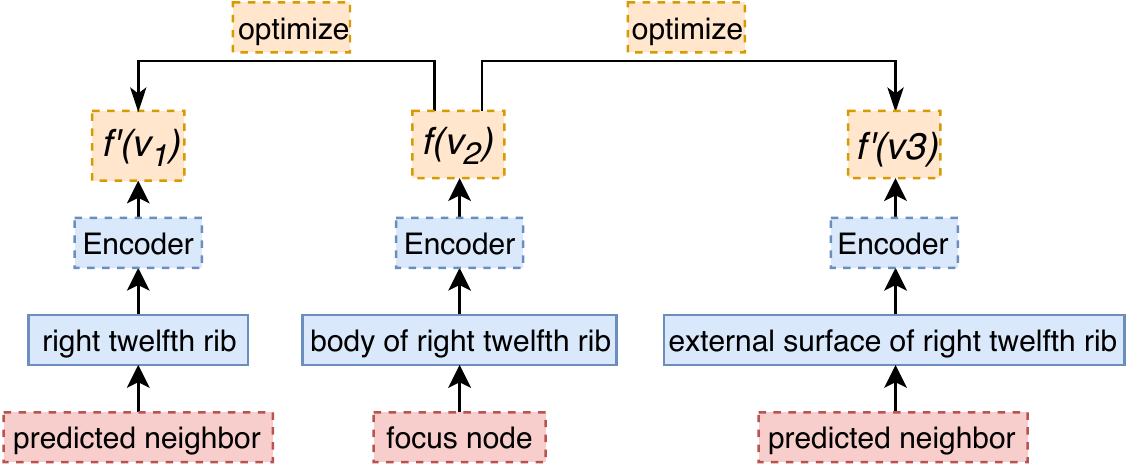}
\caption{Illustration of the proposed \nemb approach.}
\vspace{-4mm}
\label{fig:framework}
\end{figure}

More formally, let every word $w \in W$ have two embeddings $\we{w}$ and $\wep{w}$, used when $w$ occurs in the descriptor of a focus node, and when $w$ occurs in the descriptor of a neighbor of a focus node (in a random walk), respectively. For every node $v \in V$ with descriptor $S(v) = (w_1,\ldots, w_n)$, we create the sequences $T(v) = \left<\we{w_1}, \ldots, \we{w_n}\right>$ and $ T'(v) = \left<\wep{w_1}, \ldots, \wep{w_n}\right>$. We then set $f(v) = \enc(T(v))$ and $f'(v) = \enc(T'(v))$, where $\enc$ is the sequence encoder. We outline below three specific possibilities for $\enc$, though it can be any neural text encoder. Note that the embeddings $f(v)$ and $f'(v)$ of each node $v$ are constructed from the word embeddings $T(v)$ and $T'(v)$, respectively, of its descriptor $S(v)$ by the encoder $\enc$. The word embeddings of the descriptor and the parameters of $\enc$, however, are also optimized during back-propagation, so that the resulting node embeddings will predict (Eq.~\ref{eq:softmax}) the actual neighbors of each focus node (Fig.~\ref{fig:framework}). For simplicity, we only mention $f(v)$ and $T(v)$ below, but the same applies to $f'(v)$ and $T'(v)$.

\smallskip
\noindent\textbf{\avgnv}:
For every node $v \in V$, this model constructs the node's embedding $f(v)$ by simply averaging the word embeddings $T(v) = \left<\we{w_1}, \ldots, \we{w_n}\right>$ of 
$S(v) = (w_{1},w_{2},\ldots,w_{n})$. 
\begin{equation}
    f(v) = \dfrac{1}{n} \sum_{i=1}^{n} \we{w_i}
\end{equation}


\smallskip
\noindent\textbf{\grunv}: Although averaging word embeddings is effective in text categorization~\cite{joulin}, it ignores word order. To account for order, we apply \rnn{s} with \gru cells \cite{gru} instead. For each node $v \in V$ with descriptor $S(v) = \left<w_{1},\dots,w_{n}\right>$, this method computes $n$ hidden state vectors $H=\left<h_{1},\dots,h_{n}\right> = {\gru}(\we{w_1}, \dots, \we{w_n})$. The last hidden state vector $h_{n}$ is the node embedding $f(v)$.

\begin{figure}[t]
\centering
\includegraphics[width=0.4\textwidth]{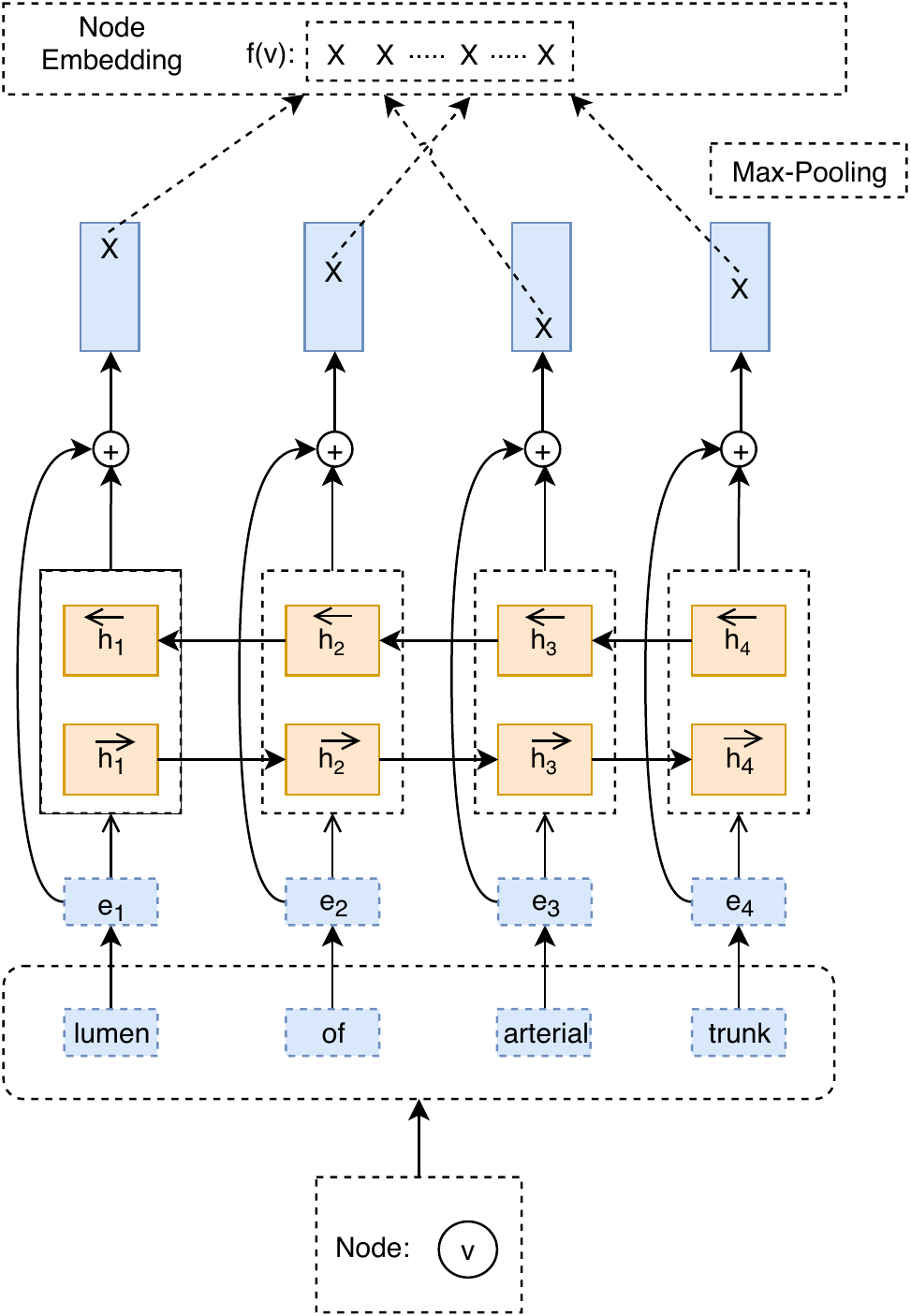}
\caption{Obtaining the embedding of a node $v$ by applying a \bigru encoder with max-pooling and 
residuals to the embeddings of $v$'s textual descriptor.}
\vspace*{-4mm}
\label{encoding2.5}
\end{figure}

\smallskip
\noindent\textbf{\bigrunv}: This method uses a bidirectional \rnn~\cite{birnn}. For each node $v$ with descriptor $S(v) = \left<w_{1},w_{2},\ldots,w_{n}\right>$, a bidirectional \gru (\bigru) computes two sets of $n$ hidden state vectors, one for each direction. These two sets are then added to form the output $H$ of the \bigru:  
\begin{eqnarray}
    {H_{f}} & = &  {\gru_{f}}(\we{w_{1}},\dots,\we{w_{n}}) \\
    {H_{b}}  & = & {\gru_{b}}(\we{w_{1}},\dots,\we{w_{n}})  \\
    H & = & {H_{f}} + {H_{b}}
\end{eqnarray}
where $_f$, $_b$ denote the forward and backward directions, and $+$ indicates component-wise addition. We add residual connections \cite{residual} from each word embedding $\we{w_t}$ to the corresponding hidden state $h_t$ of $H$. 
Instead of using the final forward and backward states of $H$, we apply max-pooling \cite{max_pooling,infersent} over the state vectors $h_t$ of $H$. The output of the max pooling is the node embedding $f(v)$. Figure~\ref{encoding2.5} illustrates this method.

Additional experiments were conducted with several variants of the last encoder. A unidirectional \gru instead of a \bigru, and a \bigru with self-attention~\cite{self_attention} instead of max-pooling were also tried. To save space, we described only the best performing variant.


\section{Experiments}

We investigate the effectiveness of our proposed approach by conducting link prediction experiments on two biomedical datasets derived from \umls. Furthermore, we devise a new approach of generating negative edges for the link prediction evaluation -- beyond just random negatives -- that makes the problem more difficult and aligns more with real-world use-cases. We also conduct a qualitative analysis, showing that the proposed framework does indeed leverage both the textual descriptors and the network structure. 


\subsection{Datasets}

We created our datasets from the \umls ontology, which contains approx.\ 3.8 million biomedical concepts and 54 semantic relationships. The relationships become edges in the networks, and the concepts become nodes. Each concept (node) is associated with a textual descriptor. We extract two types of semantic relationships, creating two networks. The first, and smaller one, consists of \partof relationships where each node represents a part of the human body. The second network contains \isa relationships, and the concepts represented by the nodes vary across the spectrum of biomedical entities (diseases, proteins, genes, etc.). To our knowledge, the \isa network is one of the largest datasets employed for link prediction and learning network embeddings. Statistics for the two datasets are shown in Table~\ref{tab:data_statistics}.

\begin{table}[!t]
\centering 
{\small 
\begin{tabular}{l|c|c}
\hline
\textbf{Statistics}   & \textbf{\isa}    & \textbf{\partof} \\
\hline
Nodes & 294,693 & 16,894  \\
\hline
Edges    & 356,541 & 19,436  \\
\hline
Training true positive edges & 294,692 & 16,893 \\
Training true negative edges & 294,692 & 16,893 \\
\hline
Test true positive edges & 61,849 & 2,543 \\
Test true negative edges & 61,849 & 2,543 \\
\hline
Avg.\ descriptor length    & 5 words      & 6  words \\
\hline
Max.\ descriptor length    & 31 words     & 14 words \\
\hline
\end{tabular}
}
\caption{Statistics of the two datasets (\isa, \partof). The true positive and true negative edges are used in the link prediction experiments.}
\vspace*{-5mm}
\label{tab:data_statistics}
\end{table}


\subsection{Baseline Node Embedding Methods}

We compare our proposed methods to baselines of two types: \emph{structure-oriented} methods, which solely focus on network structure, and \emph{content-oriented} methods that try to combine the network structure with the textual descriptors of the nodes (albeit using impoverished network neighborhoods so far). For the first type of methods, we employ \nodetovec \cite{node2vec}, which uses a biased random walk algorithm based on \deepwalk \cite{deepwalk} to explore the structure of the network more efficiently. Our work can be seen as an extension of \nodetovec that incorporates textual node descriptors, as already discussed, hence it is natural to compare to \nodetovec. As a \emph{content-oriented} baseline we use \cane~\cite{cane}, which learns separate text-based and network-based embeddings, and uses a mutual attention mechanism to dynamically change the text-based embeddings for different neighbors (Section~\ref{sec:relatedWork}). \cane only considers the direct neighbors of each node, unlike \nodetovec, which considers larger neighborhoods obtained via random walks.


\subsection{Link Prediction}

In link prediction, we are given a network with a certain fraction of edges removed. We 
need to infer these missing edges by observing the incomplete network, facilitating the discovery of links (e.g., unobserved protein-protein interactions).

Concretely, given a network we first randomly remove some percentage of edges, 
ensuring that the network remains connected so that we can perform random walks over it. Each removed edge $e$ connecting nodes $v_1, v_2$ is treated as a \emph{true positive}, in the sense that a link prediction method should infer that an edge should be added between $v_1, v_2$. We also use an equal number of \emph{true negatives}, which are pairs of nodes $v'_1, v'_2$ with no edge between $v'_1, v'_2$ in the original network. When evaluating \nemb methods, a link predictor is given true positive and true negative pairs of nodes, and is required to discriminate between the two classes by examining only the node embeddings of each pair. Node embeddings are obtained by applying a \nemb method to the pruned network, i.e., after removing the true positives. A \nemb method is considered better than another one, if it leads to better performance of the same link predictor.  

We experiment with two approaches to obtain true negatives. In \emph{Random Negative Sampling}, we randomly select pairs of nodes that were not directly connected (by a single edge) in the original network. In \emph{Close Proximity Negative Sampling}, we iterate over the nodes of the original network considering each one as a focus. For each focus node $v$, we want to find another node $u$ in close proximity that is not an ancestor or descendent (e.g., parent, grandparent, child, grandchild) of $v$ in the \isa or \partof hierarchy, depending on the dataset. We want $u$ to be close to $v$, to make it more difficult for the link predictor to infer that $u$ and $v$ should not be linked. We do not, however, want $u$ to be an ancestor or descendent of $v$, because the \isa and \partof relationships of our datasets are transitive. For example, if $u$ is a grandparent of $v$, it could be argued that inferring that $u$ and $v$ should be linked, is \emph{not} an error. To satisfy these constraints, we first find the ancestors of $v$ that are between 2 to 5 hops away from $v$ in the original network.\footnote{The edges of the resulting datasets are not directed. Hence, looking for descendents would be equivalent.} We randomly select one of these ancestors, and then we randomly select as $u$ one of the ancestor's children in the original network, ensuring that $u$ was not an ancestor or descendent of $v$ in the original network. In both approaches, we randomly select as many true negatives as the true positives, discarding the remaining true negatives. 

\smallskip
\noindent We experimented with two link predictors:

\smallskip
\noindent\textbf{Cosine similarity link predictor (\CS)}: Given a pair of nodes $v_1, v_2$ (true positive or true negative edge), \CS computes the cosine similarity (ignoring negative scores) between the two node embeddings as $s(v_1, v_2) = \max(0, \cos(f(v_1), f(v_2)))$, and predicts an edge between the two nodes if $s(v_1, v_2) \geq t$, where $t$ is a threshold. We evaluate the predictor on the true positives and true negatives (shown as `test' true positives and `test' true negatives in Table~\ref{tab:data_statistics}) by computing \auc (area under \roc curve), in effect considering the precision and recall of the predictor for varying $t$.\footnote{We do not report precision, recall, \fone scores, because these require selecting a particular threshold $t$ values.} 

\smallskip
\noindent\textbf{Logistic regression link predictor (\LR)}: Given a pair of nodes $v_1, v_2$, \LR computes the Hadamard (element-wise) product of the two node embeddings $f(v_1) \odot f(v_2)$ and feeds it to a logistic regression classifier to obtain a probability estimate $p$ that the two nodes should be linked. The predictor predicts an edge between $v_1, v_2$ if $p \geq t$. We compute \auc on a test set by varying $t$. The test set of this predictor is the same set of true positives and true negatives (with Random or Close Proximity Negative Sampling) that we use when evaluating the \CS predictor. The training set of the logistic regression classifier contains as true positives all the other edges of the network that remain after the true positives of the test set have been removed, and an equal number of true negatives (with the same negative sampling method as in the test set) that are not used in the test set. 


\subsection{Implementation Details}

For \nodetovec and our \nemb methods, which can be viewed as extensions of \nodetovec, the dimensionality of the node embeddings is 30. The dimensionality of the word embeddings (in our \nemb methods) is also 30. In the random walks, we set $r$ = 5, $l$ = 40, $k$ = 10 for \isa, and $r$ = 10, $l$ = 40, $k$ = 10 for \partof; these hyper-parameters were not particularly tuned, and their values were selected mostly to speed up the experiments. We train for one epoch with a batch size of 128, setting the number of \skipgram's negative samples to 2. We use the Adam~\cite{adam} optimizer in our \nemb methods. We implemented our \nemb methods and the two link predictors using PyTorch~\cite{pytorch} and Scikit-Learn~\cite{scikit}. For \nodetovec and \cane, we used the implementations provided.\footnote{See \url{https://github.com/aditya-grover/node2vec}, \url{https://github.com/thunlp/CANE}. 
}

For \cane, we set the dimensionality of the node embeddings to 200, as in the 
work of \citet{cane}. We also tried 30-dimensional node embeddings, as in \nodetovec and our \nemb methods, but performance deteriorated significantly. 


\subsection{Link Prediction Results}

Link prediction results for the \isa and \partof networks are reported in Tables \ref{tab:auc_isa} and \ref{tab:auc_partof}. All content-oriented \nemb methods (\cane and our extensions of \nodetovec) clearly outperform the structure-oriented method (\nodetovec) on both datasets in both negative edge sampling settings, showing that modeling the textual descriptors of the nodes is critical. Furthermore, all methods perform much worse with Close Proximity Negative Sampling, confirming that the latter produces more difficult link prediction datasets.

All of our \nemb methods (content-aware extensions of \nodetovec) outperform \nodetovec and \cane in every case, especially with Close Proximity Negative Sampling. We conclude that it is important to model not just the textual descriptor of a node or its direct neighbors,
but as much non-local network structure as possible.

\begin{table}[!t]\small
\begin{center}
    \begin{tabular}{@{}l|c|c@{}}
        \toprule
         &  \textbf{Random} & \textbf{Close}  \\ 
         & \textbf{Negative}  &  \textbf{Proximity}  \\ 
        \textbf{NE Method + Link Predictor} &  
        \textbf{Sampling} & \textbf{Sampling} \\ 
        \midrule
        {Node2Vec + CS} & 66.6 & 54.3 \\
        {CANE + CS} & 94.1  & 69.6 \\
        {Avg-N2V + CS} & 95.0 & 78.6 \\
        {GRU-N2V + CS} & \textbf{98.7} & \textbf{79.2} \\
        {BiGRU-Max-Res-N2V + CS} & 98.5 & 79.0 \\
        \midrule
        \begin{tabular}[c]{@{}l@{}}
        {Node2Vec + LR}\\
        {CANE + LR}\\
        {Avg-N2V + LR}\\ 
        {GRU-N2V + LR}\\ 
        {BiGRU-Max-Res-N2V + LR}\end{tabular} &
        \begin{tabular}[c]{@{}l@{}}77.2\\95.3 \\ 97.6\\ 99.0\\ \textbf{99.3}\end{tabular} & \begin{tabular}[c]{@{}l@{}}56.3\\70.0 \\73.9\\ 79.6\\ \textbf{82.1}\end{tabular} \\ \bottomrule
    \end{tabular}
    \caption{\auc scores (\%) for the \isa dataset. Best scores per link predictor (\CS, \LR) shown in bold.}
    \vspace*{-2mm}
    \label{tab:auc_isa}
\end{center}
\end{table}

\begin{table}[!t]\small
\begin{center}
    \begin{tabular}{@{}l|c|c@{}}
        \toprule
         &  \textbf{Random} & \textbf{Close}  \\ 
         & \textbf{Negative}  &  \textbf{Proximity}  \\ 
        \textbf{NE Method + Link Predictor} &  
        \textbf{Sampling} & \textbf{Sampling} \\ 
        \midrule
        {Node2Vec + CS} & 76.8 & 61.8 \\
        {CANE + CS} & 93.9  & 75.3 \\
        {Avg-N2V + CS} & 95.9  & 81.8 \\
        {GRU-N2V + CS} & 98.0 & 83.1 \\
        {BiGRU-Max-Res-N2V + CS} & \textbf{98.5} & \textbf{83.3} \\
        \midrule
        \begin{tabular}[c]{@{}l@{}}
        {Node2Vec + LR} \\
        {CANE + LR} \\
        {Avg-N2V + LR}\\ 
        {GRU-N2V + LR}\\ 
        {BiGRU-Max-Res-N2V + LR}\end{tabular} &
        \begin{tabular}[c]{@{}l@{}}85.2\\ 94.4\\ 97.6\\ 99.0\\ \textbf{99.5}\end{tabular} & \begin{tabular}[c]{@{}l@{}}66.5\\ 76.3\\ 79.4\\ 85.6\\ \textbf{88.6}\end{tabular} \\ \bottomrule
    \end{tabular}
    \caption{\auc scores (\%) for the \partof dataset. Best scores per link predictor (\CS, \LR) shown in bold.}
    \vspace*{-4mm}
    \label{tab:auc_partof}
\end{center}
\end{table}

For \partof relations (Table~\ref{tab:auc_partof}), \bigrunv obtains the best results with both link predictors (\CS, \LR) in both negative sampling settings, but the differences from \grunv are very small in most cases. For \isa (Table~\ref{tab:auc_isa}), \bigrunv obtains the best results with the \LR predictor, and only slightly inferior results than \grunv with the \CS predictor. The differences of these two \nemb methods from \avgnv are larger, indicating that recurrent neural encoders of textual descriptors are more effective than simply averaging the word embeddings of the descriptors.

Finally, we observe that the best results of the \LR predictor are better than those of the \CS predictor, in both datasets and with both negative edge sampling approaches, with the differences being larger with Close Proximity Sampling. This is as one would expect, because the logistic regression classifier can assign different weights to the dimensions of the node embeddings, depending on their predictive power, whereas cosine similarity assigns the same importance to all dimensions. 

\begin{table}[!t]\small
\begin{center}
    \begin{tabular}{@{}l|l|c@{}}
        \toprule
        \multicolumn{3}{c}{\textbf{Target Node:} Left Eyeball (\partof)} \\ \midrule
        \textbf{Most Similar Embeddings}                                 & \textbf{Cos}    &\textbf{Hops}\\ \midrule
        equator of left eyeball              & 99.3    & 1 \\ \midrule
        episcleral layer of left eyeball     & 99.2    & 4 \\ \midrule
        cavity of left eyeball               & 99.1    & 1\\ \midrule
        wall of left eyeball                 & 99.0    & 1\\ \midrule
        vascular layer of left eyeball       & 98.9    & 1\\
    \end{tabular}
    \bigskip
    \begin{tabular}{@{}l|l|c@{}}
        \toprule
        \multicolumn{3}{c}{\textbf{Target Node:} Lung Carcinoma (\isa)}                               \\ \midrule
        \textbf{Most Similar Embeddings}                                                                            & \textbf{Cos} & \textbf{Hops}\\ \midrule
        \begin{tabular}[c]{@{}l@{}}recurrent lung \\ carcinoma\end{tabular}             & 97.6  & 1\\ \midrule
        papillary carcinoma                                                             & 97.1  & 2\\ \midrule
        \begin{tabular}[c]{@{}l@{}}lung pleomorphic \\ carcinoma\end{tabular}           & 97.0  & 3\\ \midrule
        ureter carcinoma                                                                & 96.6  & 2\\ \midrule
        \begin{tabular}[c]{@{}l@{}}lymphoepithelioma-like lung\\ carcinoma\end{tabular} & 96.6  & 3\\ \bottomrule
    \end{tabular}
    \vspace*{-4mm}
    \caption{Examples of nodes whose embeddings are closest (cosine similarity, Cos) to the embedding of a target node in the \partof (top) and \isa (bottom) datasets. We also show the distances (number of edges, Hops) between the nodes in the networks.}
    \vspace*{-4mm}
    \label{tab:mostSimilar}
\end{center}
\end{table}


\subsection{Qualitative Analysis}

To better understand the benefits of leveraging both network structure and textual descriptors, we present examples from the two datasets.

\smallskip
\noindent \textbf{Most similar embeddings}: Table~\ref{tab:mostSimilar} presents the five nearest nodes for two target nodes (`Left Eyeball' and `Lung Carcinoma'), based on the cosine similarity of the corresponding node embeddings in the \partof and \isa networks, respectively. We observe that all nodes in the \partof example are very similar content-wise to our target node. Furthermore, the model captures the semantic relationship between concepts, since most of the returned nodes are actually parts of `Left Eyeball'. The same pattern is observed in the \isa example, with the exception of `ureter carcinoma', which is not directly related with `lung carcinoma', but is still a form of cancer. Finally, it is clear that the model extracts meaningful information from both the textual content of each node and the network structure, since the returned nodes are closely located in the network (Hops 1--4). 

\begin{figure}
\vspace*{-2mm}
	\centering
	\begin{subfigure}{0.4\textwidth}
		\includegraphics[width=0.94\textwidth]{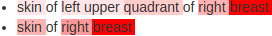}
		\caption{Two nodes connected by a \partof edge.}
	\end{subfigure}
	\smallskip
	\begin{subfigure}{0.4\textwidth}
		\includegraphics[width=0.82\textwidth]{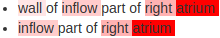}
		\caption{Two nodes connected by a \partof edge.}
	\end{subfigure}
    \smallskip
    \begin{subfigure}{0.39\textwidth}
		\includegraphics[width=0.85\textwidth]{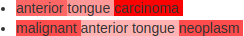}
		\caption{Two nodes connected by an \isa edge.}
	\end{subfigure}
	\vspace*{-2mm}
	\caption{Visualization of the importance that \bigrunv assigns to the words of the descriptors of the nodes of three edges. Edges (a) and (b) are from the \partof dataset. Edge (c) is from the \isa dataset.}
	\label{fig:heatmap}
\end{figure}

\begin{table}[!htb]\small
    \begin{tabular}{l|c|c|c|c}
        \toprule
        \textbf{Edges/Descriptors} & \textbf{BN2V} & \textbf{CANE} & \textbf{N2V} & \textbf{Hops} \\
        \midrule
        \begin{tabular}[c]{@{}l@{}}
        (a) bariatric surgery\\ (b) bypass \\ gastrojejunostomy
        \end{tabular}  
        & 82.7   & 38.0 & 56.2   & 11   \\
        \midrule
        \begin{tabular}[c]{@{}l@{}}
        (a) anatomical line\\ 
        (b) anterior \\ malleolar fold
        \end{tabular}             
        & 82.3   & 29.0 & 50.0   & 22   \\
        \midrule
        \begin{tabular}[c]{@{}l@{}}
        (a) zone of biceps \\ brachii\\ 
        (b) short head of \\ biceps brachii 
        \end{tabular} 
        & 93.0   & 70.0 & 61.6   & 13 \\
        \bottomrule
    \end{tabular}
    \caption{Examples of true positive edges, showing how structure and textual descriptors affect node embeddings. The first two edges are \isa, the third one is \partof. The \nemb methods used are \bigrunv (\textsc{bn2v}), \cane and \nodetovec (\textsc{n2v}). We report cosine similarities between node embeddings and the distances between the nodes (number of edges, Hops) in the networks after removing true positive edges.}
    \vspace*{-5mm}
    \label{tab:casestudy}
\end{table}

\vspace{2mm}
\noindent \textbf{Heatmap visualization}: \bigrunv can be extended to highlight the words in each textual descriptor that mostly influence the corresponding node embedding. Recall that this \nemb method applies a max-pooling operator (Fig.~\ref{encoding2.5}) over the state vectors $h_1, \dots, h_n$ of the words $w_1, \dots, w_n$ of the descriptor, keeping the maximum value per dimension across the state vectors. We count how many dimension-values the max-pooling operator keeps from each state vector $h_i$, and we treat that count (normalized to $[0,1]$) as the importance score of the corresponding word $w_i$.\footnote{We actually obtain two importance scores for each word in the descriptor of a node, since each node $v$ has two embeddings $f(v), f'(v)$, used when $v$ is the focus or a neighbor (Section \ref{sec:methods}), i.e., there are two results of the max-pooling operator. We average the two importance scores of each word.} We then visualize the importance scores as heatmaps of the descriptors. In the first two example edges of  Fig.~\ref{fig:heatmap}, the highest importance scores are assigned to words indicating body parts, which is appropriate given that the edges indicate \partof relations. In the third example edge, the highest importance score of the first descriptor is assigned to `carcinoma', and the highest importance scores of the second descriptor are shared by `malignant' and `neoplasm'; again, this is appropriate, since these words indicate an \isa relation.



\vspace{2mm}
\noindent {\bf Case Study}: In Table~\ref{tab:casestudy}, we present examples that illustrate learning from both the network structure and textual descriptors. All three edges are true positives, i.e., they were initially present in the network and they were removed to test link prediction. In the first two edges, which come from the \isa network, the node descriptors share no words. Nevertheless, \bigrunv (\textsc{bn2v}) produces node embeddings with high cosine similarities, much higher than \nodetovec that uses only network structure, presumably because the word embeddings (and neural encoder) of \textsc{bn2v} correctly capture lexical relations (e.g., near-synonyms). Although \cane also considers the textual descriptors, its similarity scores are much lower, presumably because it uses only local neighborhoods (single-edge hops). The nodes in the third example, which come from the \partof network, have a larger word overlap. \nodetovec is unaware of this overlap and produces the lowest score. The two content-oriented methods (\textsc{bn2v}, \cane) produce higher scores, but again \textsc{bn2v} produces a much higher similarity, presumably because it uses larger neighborhoods. In all three edges, the two nodes are distant ($>$10 hops), yet \textsc{bn2v} produces high similarity scores.


\section{Conclusions and Future Work}

We proposed a new method to learn content-aware node embeddings, which extends \nodetovec by considering the textual descriptors of the nodes. The proposed approach leverages the strengths of both structure- and content-oriented node embedding methods. It exploits non-local network neighborhoods generated by random walks, as in the original \nodetovec, and allows integrating various neural encoders of the textual descriptors. We evaluated our models on two biomedical networks extracted from \umls, which consist of \partof and \isa edges. Experimental results with two link predictors, cosine similarity and logistic regression, demonstrated that our approach is effective and outperforms previous methods which rely on structure alone, or model content along with local network context only. 

In future work, we plan to experiment with networks extracted from other biomedical ontologies and knowledge bases. We also plan to explore if the word embeddings that our methods generate can improve biomedical question answering systems \cite{McDonald2018}.


\section*{Acknowledgements}
This work was partly supported by the Research Center of the Athens University of Economics and Business. The  work was also  supported by the French National Research Agency under project ANR-16-CE33-0013.

\bibliographystyle{acl_natbib}
\bibliography{node2vec}

\pagebreak
\newpage
\appendix
\section*{Appendix}
\section{CANE Hyper-parameters}
\cane 
has 3 hyper-parameters, denoted $\alpha$, $\beta$, $\gamma$, 
which control to what extent it uses information from network structure or textual descriptors.
We learned these hyper-parameters by employing the HyperOpt~\cite{HyperoptAP} 
on the validation set.\footnote{ 
For more information on HyperOpt see: \url{https://github.com/hyperopt/hyperopt/wiki/FMin}.
For a tutorial see: \url{https://github.com/Vooban/Hyperopt-Keras-CNN-CIFAR-100}.}
All three hyper-parameters had the same search space: 
$[0.2, 1,0]$ with a step of $0.1$. The optimization 
ran for 30 trials for both datasets. Table~\ref{tab:params} reports the resulting hyper-parameter values. 

\begin{table}[!htb]\small
\centering
\begin{tabular}{@{}l|l|l@{}}
\toprule
\textbf{Parameters} & \textbf{\partof} & \textbf{\isa} \\ \midrule
$\alpha$          & 0.2     & 0.7 \\ \midrule
$\beta$          & 1.0     & 0.7  \\ \midrule
$\gamma$          & 1.0     & 0.7 \\
\bottomrule
\end{tabular}
\caption{
Hyper-parameter values used in \cane.
}
\vspace*{-5mm}
\label{tab:params}
\end{table}

\end{document}